\documentclass[letterpaper, 10 pt, conference]{ieeeconf}  % Comment this line out if you need a4paper

\usepackage{amsmath}
\usepackage{adjustbox}
\usepackage{graphicx,float}
\usepackage{epstopdf}
\usepackage{caption}
\usepackage{pdflscape}
\usepackage{amssymb}
\usepackage{amsfonts}
\usepackage{pgfplots}
\usepackage{bbm}
\usepackage{color}
\usepackage{multirow}
\usepackage{tikz}
\usepackage{dblfloatfix}
\usepackage{float}
\usepackage{hyperref}
\usepackage{adjustbox}
\usepackage{balance}
\usepackage{algorithm}
\usepackage{algpseudocode}
\usepackage{relsize}
\usepackage{subcaption}
\usepackage{tabularx} % Add this to your preamble
\usepackage[compress]{cite}
\usepackage{arydshln}
\usetikzlibrary{arrows.meta,positioning,calc,shapes.geometric}

\usepackage{verbatim}  % Add this in the preamble

\makeatletter
\newcommand*\titleheader[1]{\gdef\@titleheader{#1}}
\AtBeginDocument{%
	\let\st@red@title\@title
	\def\@title{%
		\bgroup\normalfont\large\centering\@titleheader\par\egroup
		\vskip1.5em\st@red@title}
}
\makeatother

\IEEEoverridecommandlockouts                              % This command is only needed if 
\titleheader{This work has been submitted to the IEEE for possible publication. Copyright may be transferred without notice, after which this version may no longer be accessible.}

\title{\LARGE \bf
Radar-Inertial Odometry with Online Spatio-Temporal Calibration via Continuous-Time IMU Modeling
}

\author{Vlaho-Josip Štironja$^{1}$, Luka Petrović$^{1}$, Juraj Peršić$^{2}$, Ivan Marković$^{1}$,  and Ivan Petrović$^{1}$% <-this % stops a space
\thanks{$^{1}$University of Zagreb Faculty of Electrical Engineering and Computing, Laboratory for Autonomous Systems and Mobile Robotics (LAMOR), Unska 3, HR-10000, Zagreb, Croatia
        {\tt\small \{name.surname\}@fer.unizg.hr}}%
\thanks{$^{2}$Calirad d.o.o.,
Augusta Harambašića 4, HR-10000, Zagreb, Croatia
        {\tt\small  juraj.persic@calirad.net}}%
}

\begin{document}

\maketitle
\thispagestyle{empty}
\pagestyle{empty}

%%%%%%%%%%%%%%%%%%%%%%%%%%%%%%%%%%%%%%%%%%%%%%%%%%%%%%%%%%%%%%%%%%%%%%%%%%%%%%%%
\begin{abstract}
Radar-Inertial Odometry (RIO) has emerged as a robust alternative to vision- and LiDAR-based odometry in challenging conditions such as low light, fog, featureless environments, or in adverse weather.
However, many existing RIO approaches assume known radar-IMU extrinsic calibration or rely on sufficient motion excitation for online extrinsic estimation, while temporal misalignment between sensors is often neglected or treated independently.
In this work, we present a RIO framework that performs joint online spatial and temporal calibration within a factor-graph optimization formulation, based on continuous-time  modeling of inertial measurements using uniform cubic B-splines.
The proposed continuous-time representation of acceleration and angular velocity accurately captures the asynchronous nature of radar-IMU measurements, enabling reliable convergence of both the temporal offset and extrinsic calibration parameters, without relying on scan matching, target tracking, or environment-specific assumptions.
Project website: \href{https://unizgfer-lamor.github.io/lc-rio-et/}{https://unizgfer-lamor.github.io/lc-rio-et/}.
\end{abstract}

%%%%%%%%%%%%%%%%%%%%%%%%%%%%%%%%%%%%%%%%%%%%%%%%%%%%%%%%%%%%%%%%%%%%%%%%%%%%%%%%
\section{Introduction}

Reliable and accurate state estimation is a fundamental requirement for autonomous systems operating in real-world environments.
While Global Navigation Satellite Systems (GNSS) provide reliable positioning in open outdoor areas, their performance degrades severely or fails entirely in indoor environments, urban canyons, and tunnels due to signal blockage and signal unreliability to robustly estimate location \cite{review}.
To ensure reliable ego-motion estimation under such conditions, multi-sensor fusion approaches that combine inertial measurement units (IMUs) with complementary exteroceptive sensors, such as cameras, LiDAR, or automotive Frequency Modulated Continuous Wave (FMCW) radar, have been widely adopted \cite{doer_ekf,doer_online,jan_first,mfi-rik}.
Among these sensing modalities, radar has emerged as a compelling alternative to vision and LiDAR-based systems.
Unlike cameras, radar maintains robust performance in adverse weather conditions, including fog, rain, and poor illumination \cite{peng20254d}.
Compared to LiDAR, radar sensors are typically more cost-effective, offer longer sensing range in degraded visibility, and provide Doppler velocity measurements.
This capability enables accurate instantaneous ego-velocity estimation even in environments with sparse geometric structure or dynamic objects \cite{inst}.
These advantages have recently driven significant research interest in radar-based localization and odometry.

\begin{figure}[t]
    \centering
    \includegraphics[width=0.47\textwidth]{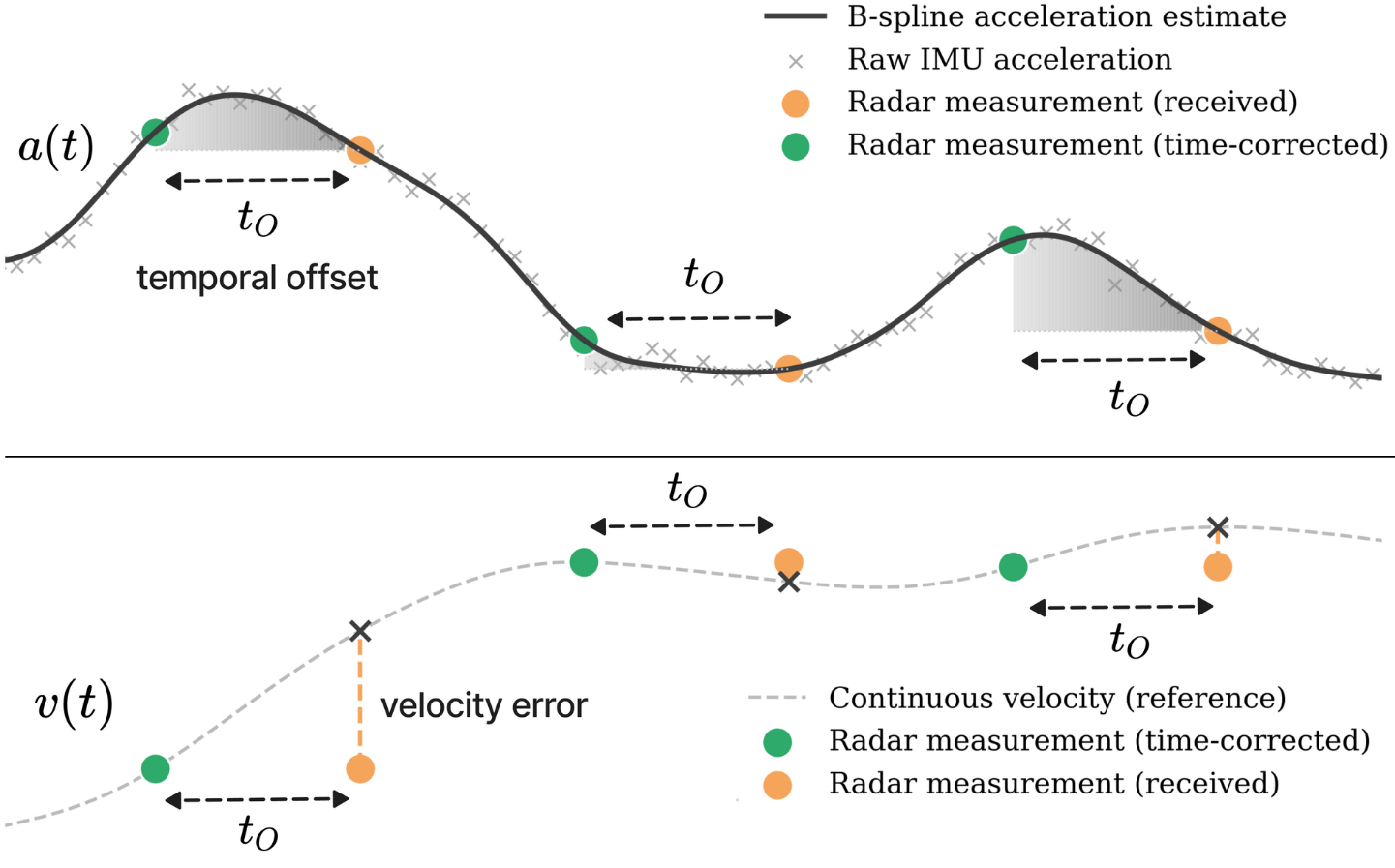}
    \caption{Temporal misalignment between discrete radar ego-velocity measurements and the continuous-time motion estimated via a uniform cubic B-spline fit to IMU acceleration signals.}
    \label{fig::first_page}
\end{figure}

Despite its robustness, radar sensing poses several challenges, including high measurement noise, limited angular resolution, and pronounced clutter and multipath reflection effects.
These characteristics make data association unreliable and error-prone, while ego-velocity estimation remains robust and accurate.
In contrast, IMUs provide high-rate and locally accurate acceleration and angular velocity measurements but suffer from drift due to sensor biases.
When fused together, radar and IMU measurements exhibit strong complementarity, the IMU supports short-term motion propagation and interpolation between radar updates, while radar provides drift-correcting constraints in environments where vision or LiDAR-based systems may fail.

Radar–Inertial Odometry has been explored using both the extended Kalman filter (EKF) \cite{doer_ekf,doer_online,dero} and factor graph optimization (FGO) frameworks \cite{mfi-rik,riv,yang2025ground}.
EKF-based approaches offer real-time performance but rely on local linearization and may suffer from reduced consistency over long trajectories.
In contrast, FGO-based methods enable more accurate and consistent estimation by jointly optimizing over a window of states and measurements.
Beyond estimator choice, prior work has improved RIO performance through enhanced radar data utilization, including point tracking constraints \cite{jan_multi}, scan registration methods \cite{amodeo20244d}, or environment-specific assumptions such as ground plane estimation \cite{yang2025ground} or Manhattan-world constraints \cite{doer2021yaw}.
A critical yet often overlooked challenge in practical RIO systems is online spatio-temporal misalignment between the radar and IMU (Fig.~\ref{fig::first_page}).
While several methods address online spatial (extrinsic) calibration \cite{jan_online,doer_online}, temporal misalignment is frequently assumed negligible or handled separately.
Recent works have begun to estimate radar–IMU time offsets online, but typically assume known and fixed extrinsics \cite{kim2025ekf,vstironja2025impact}.

In this paper, we propose LC-RIO-ET (Loosely Coupled RIO with online Extrinsic and Temporal calibration), a unified factor graph optimization framework for RIO with joint online spatial and temporal calibration, enabling simultaneous online estimation of both calibration parameters alongside the system state, without relying on scan matching, target tracking, or environment-specific assumptions. 
At the core of the framework is a novel radar ego-velocity factor formulation based on continuous-time inertial modeling using uniform cubic B-splines, allowing acceleration and angular velocity to be evaluated at arbitrary timestamps with $C^2$-continuous derivatives.
To the best of our knowledge, no existing work performs RIO with joint online spatial and temporal calibration within a unified factor graph framework. 
Furthermore, we extensively evaluate the proposed method on real-world radar–IMU datasets in both hardware-synchronized and non-synchronized configurations, demonstrating reliable calibration convergence and improved odometry accuracy.

\section{Related Work}
This section reviews RIO approaches based on automotive FMCW 4D radars, which measure range, azimuth, elevation, and Doppler velocity, focusing on estimation methods and online spatio-temporal calibration.

\subsection{Radar-Inertial Odometry} 
Most RIO approaches integrate ego-velocity estimates from Doppler radar measurements under the assumption that the observed environment is static. 
As this assumption is frequently violated in practice, robust outlier rejection is essential, with RANSAC being one of the most commonly adopted strategies.
Kellner et al.~\cite{inst} introduced instantaneous ego-velocity estimation from Doppler measurements, which subsequently served as the foundation for several RIO systems.
To improve robustness, Zhuang et al. \cite{iriom} proposed a non-minimal solver with graduated non-convexity (GNC), which has been shown to be more resilient than RANSAC in cluttered radar data.
The \textsc{RAVE} framework \cite{rave} provides a systematic evaluation of radar ego-velocity estimators and outlier rejection strategies, further proposing a moving-average filter to discard physically infeasible velocity estimates.
Building upon the idea of instantaneous ego-velocity estimation \cite{inst}, Doer and Trommer \cite{doer_ekf} extended instantaneous Doppler-based estimation to full 3D ego-velocity and fused ego-velocity with IMU measurements in a loosely coupled error-state EKF, further incorporating barometric measurements to reduce vertical drift.
In subsequent work, they improved yaw estimation in indoor environments by exploiting Manhattan-world assumptions \cite{doer2021yaw}.

An alternative formulation, dubbed DeRO \cite{dero}, combines radar ego-velocity with gyroscope measurements, reducing sensitivity to accelerometer bias and double integration.  
Instead of relying on direct accelerometer integration, DeRO estimates tilt from gravity and incorporates radar scan-matching-derived distance constraints during the measurement update.
Girod et al. \cite{mfi-rik} introduced a factor-graph-based baro-radar-inertial odometry framework that jointly optimizes Doppler velocity, bearing constraints, robust differential barometry, and zero-velocity updates. 
Michalczyk et al. \cite{jan_first}  proposed a tightly coupled EKF-based RIO that explicitly incorporates radar range measurements, reducing reliance on angular information that is often noisy in low-cost FMCW radars. 
This framework was later extended \cite{jan_multi} by introducing persistent landmark tracking across multiple poses, an approach analogous to long-term feature tracking in visual-inertial odometry, yielding substantially reduced drift. 
More recently, the same authors explored learning-based radar point correspondence estimation using transformer architectures trained in a self-supervised manner via set-based matching losses \cite{michalczyk2025learning}.

A common extension of RIO estimators is the integration of radar scan registration.
However, Kubelka et al. \cite{do_we_need_sm} argue that Doppler-based ego-velocity fusion with IMU measurements often achieves accuracy comparable to, or exceeding, scan-to-scan and scan-to-map registration, particularly for low-cost automotive radars with sparse returns. 
Similar conclusions were drawn in \cite{movro2}, where authors demonstrate that scan matching primarily improves estimator robustness in failure cases of complementary sensors rather than consistently improving accuracy.
Recent work has proposed improved radar scan-matching techniques which could lead to more accurate constraint and overall odometry estimates. 
Amodeo et al. \cite{amodeo20244d} introduced a probabilistic registration approach based on summarized 3D Gaussian representations rather than voxelized maps, while Kim et al. \cite{kim2025doppler} proposed Doppler-based correspondences that are invariant to translation and small rotations, yielding more accurate alignment.

Several RIO systems further exploit environmental structure. 
Yang et al. \cite{yang2025ground} proposed a ground-optimized RIO framework that models ground surfaces with uncertainty-aware zone representations and performs continuous-time velocity integration using Gaussian processes to address asynchronous sensor data. 
Noh et al. \cite{noh2025garlio} used gravity and Doppler measurements to reduce vertical drift, later extending this idea \cite{noh2025garlileo} to a continuous-time gravity-aligned radar-leg-inertial odometry framework that represents ego-velocity as a spline driven by radar and kinematic measurements.

Finally, Huang et al. \cite{less_is_more} demonstrated that incorporating radar cross-section (RCS) information strengthens point filtering, correspondence estimation, and residual modeling. By explicitly modeling radar point uncertainty \cite{unc_2025} in polar coordinates, their approach better reflects radar sensing characteristics and improves estimation accuracy.

\subsection{Online Spatio-Temporal Calibration in Radar-Inertial Odometry}
Online extrinsic calibration in RIO is commonly performed by augmenting the estimator state with extrinsic calibration parameters which are assumed to remain constant over time.
Under sufficient excitation, these parameters become observable and can converge during operation, improving the accuracy of odometry and eliminating the need for manual calibration. 
This has been demonstrated in both EKF-based and optimization-based RIO frameworks \cite{doer_online,jan_online, iriom}, where online spatial calibration improves estimation accuracy under sufficient dynamic motion.

While online temporal calibration has been extensively studied in other multi-sensor fusion settings, such as visual-inertial odometry, e.g. \cite{qin2018online}, it has remained largely unexplored in RIO until recently. 
Kim et al. \cite{kim2025ekf} presented the first EKF-based RIO framework with online temporal calibration, in which the radar ego-velocity measurement model, derived from a single radar scan, explicitly incorporates a time offset parameter into the filter update. 
Their results demonstrated that accounting for temporal misalignment substantially improves the accuracy of state estimation.
Štironja et al.~\cite{vstironja2025impact} introduced a factor-graph-based RIO with online temporal calibration, correcting radar ego-velocity by assuming locally constant acceleration around the measurement time. 
While this approach converges reliably for small temporal delays, it requires multiple iterations under larger offsets.
Kelly et al. \cite{kelly2021question} have highlighted fundamental limitations of recursive filtering approaches for temporal calibration, noting that they may introduce distortions and inconsistencies even under careful tuning.

Our proposed factor graph formulation addresses these limitations, as the optimization window can explicitly span the full range of possible temporal offsets. 
In contrast to RIO-T \cite{vstironja2025impact}, which corrects radar ego-velocity by assuming locally constant acceleration around the measurement time, we model inertial measurements, both angular velocity and acceleration, as continuous-time trajectories using uniform cubic B-splines.
This allows acceleration and angular velocity to be evaluated at arbitrary timestamps with $C^2$-continuous derivatives, enabling direct integration of the temporal offset into the measurement model and yielding a more accurate representation of system dynamics, particularly under high-motion excitation and large temporal offsets.

\section{Proposed Approach}
\label{sec:method}

The proposed method LC-RIO-ET is formulated within a FGO framework.
LC-RIO-ET jointly estimates the temporal offset and extrinsic parameters between radar and inertial measurements in an online manner, without relying on environmental assumptions such as ground-plane or Manhattan world constraints.
The key contribution of this work is the use of uniform cubic B-splines to  model inertial signals in continuous time, enabling acceleration and angular velocity estimation at arbitrary timestamps.

\subsection{Notation, State Definition, and Assumptions}

Scalars are denoted by non-bold lowercase letters (e.g., $m$), vectors by  bold lowercase letters (e.g., $\mathbf{m}$), and matrices by bold uppercase  letters (e.g., $\mathbf{M}$).
Coordinate frames are denoted by $\mathcal{F}$, and transformations follow  the notation in~\cite{notation,barfoot}.
The translation from frame $\mathcal{F}_B$ to $\mathcal{F}_C$, expressed in  frame $\mathcal{F}_A$, is written as ${}_{\!A}\mathbf{t}_{BC}$, while  $\mathbf{R}_{AB} \in \mathrm{SO}(3)$ denotes the rotation from $\mathcal{F}_B$  to $\mathcal{F}_A$.
The system state is defined as:
\begin{equation}
\setlength{\arraycolsep}{3pt}
\mathbf{x} =
\begin{bmatrix}
\mathbf{R}_{\mathrm{IW}}, &
{}_{\!W}\mathbf{p}_{\mathrm{WI}}, &
{}_{\!W}\mathbf{v}, &
\mathbf{b}_{g}, &
\mathbf{b}_{a}, &
\mathbf{R}_{\mathrm{IR}}, &
{}_{\!I}\mathbf{p}_{\mathrm{IR}}, &
t_{O}
\end{bmatrix},
\label{eq:state}
\end{equation}
where $\mathbf{R}_{\mathrm{IW}}$, ${}_{\!W}\mathbf{p}_{\mathrm{WI}}$ denote  the system orientation and position, ${}_{\!W}\mathbf{v}$ is the velocity in the world frame, $\mathbf{b}_{g}$ and $\mathbf{b}_{a}$ are the gyroscope and accelerometer biases, $\mathbf{R}_{\mathrm{IR}}$ and ${}_{\!I}\mathbf{p}_{\mathrm{IR}}$ represent the radar–IMU extrinsic calibration, and $t_{O}$ is the temporal offset between radar and IMU.
The world frame $\mathcal{F}_W$ is initialized at system start, aligned 
with the initial IMU frame.

At initialization, the gyroscope bias and the roll and pitch angles are estimated from a stationary phase, while the accelerometer bias, position, velocity, and temporal offset are set to zero.
The radar–IMU extrinsic calibration is assumed to be approximately known, and the initial state covariance is treated as a tunable parameter.

\subsection{Factor Graph Formulation}

FGO for multi-sensor fusion can be posed as a maximum a posteriori (MAP) estimation problem~\cite{barfoot}, in which the system states are inferred by  jointly exploiting probabilistic measurement models and prior information~\cite{gtsam_intro}.
Under the Markov assumption and that measurements are conditionally independent, the MAP estimation problem can be written as:
\begin{equation}
\hat{\mathbf{x}}=\underset{\mathbf{x}}{\mathrm{argmax}} 
\prod_{k,i}p_i(\mathbf{z}_{k,i} \mid \mathbf{x}_k)
\prod_{k}p(\mathbf{x}_k \mid \mathbf{x}_{k-1},\mathbf{u}_k),
\label{fgo_1}
\end{equation}
where $\mathbf{x}_k$ denotes the system state at time step $k$, $\mathbf{z}_{k,i}$ represents the observation from the $i$-th sensor at time $k$, and $\mathbf{u}_k$ corresponds to the control input or motion prior.
Assuming zero-mean Gaussian noise with covariance $\Sigma_{k,i}$, the MAP estimation reduces to a nonlinear least-squares problem:
\begin{equation}
\hat{\mathbf{x}} = \underset{\mathbf{x}}{\mathrm{argmin}} 
\sum_{k,i}\left| \mathbf{z}_{k,i} - 
\mathbf{h}_{k,i}(\mathbf{x}_k) \right|^2_{\Sigma_{k,i}},
\label{fgo_4}
\end{equation}
where $\mathbf{h}_{k,i}(\cdot)$ denotes the measurement model associated with the $i$-th sensor at time $k$.

\begin{figure}[t]
\vspace{5mm}
\centering
\includegraphics[width=0.45\textwidth]{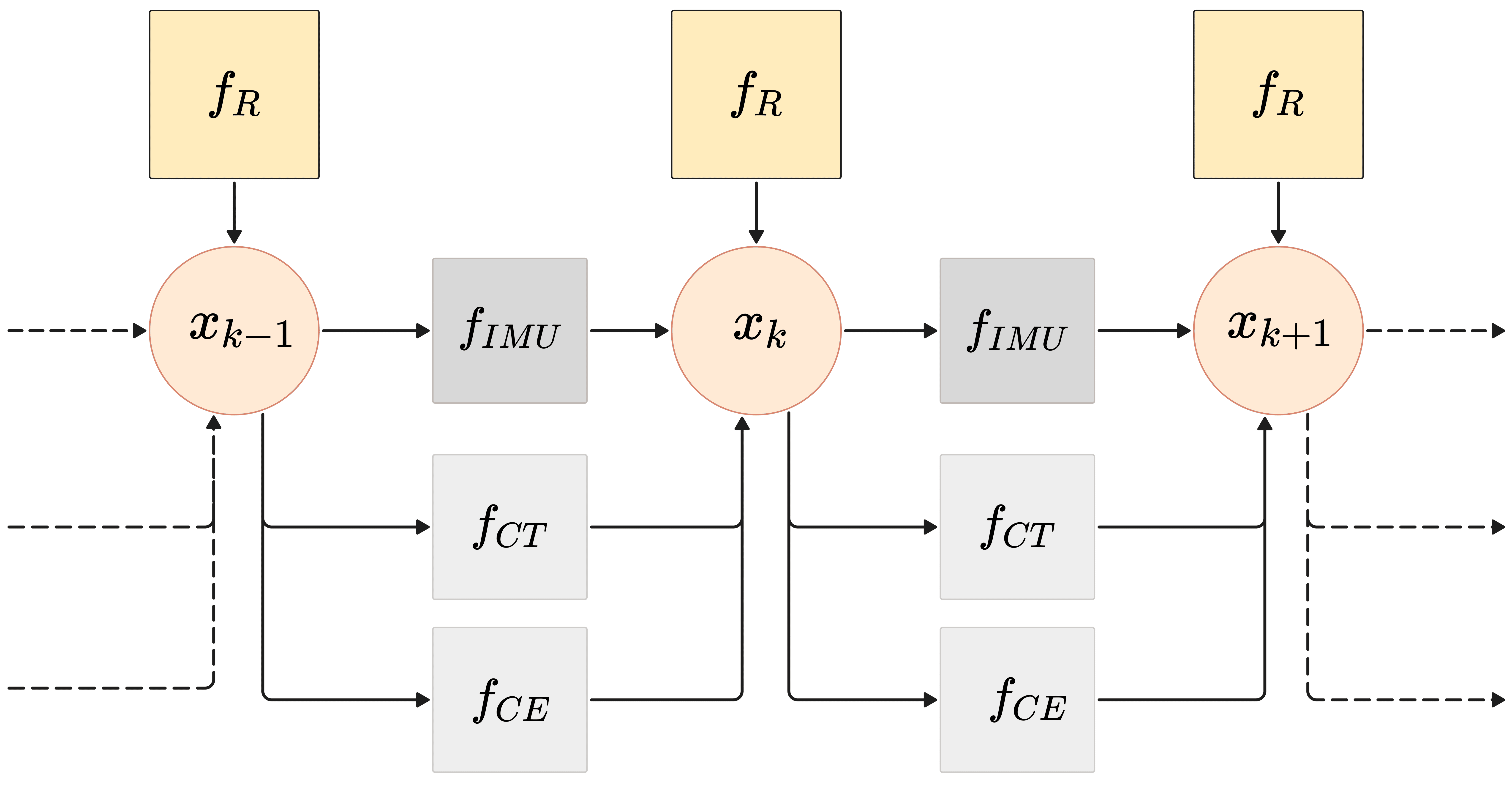}
\caption{Factor graph architecture employed in this work. The graph comprises an IMU factor ($f_{IMU}$), a radar ego-velocity factor ($f_{R}$), and a constant time-offset ($f_{CT}$) and constant extrinsic calibration factors ($f_{CE}$).}
\label{fig::fgo_structure}
\vspace{-5mm}
\end{figure}

The proposed factor graph structure is illustrated in Fig.~\ref{fig::fgo_structure}. The formulation comprises four factors: (1) the IMU preintegration factor ($f_{\mathrm{IMU}}$), (2) the radar ego-velocity factor ($f_R$), (3) the temporal offset consistency factor ($f_{\mathrm{CT}}$), and (4) the extrinsic calibration consistency factor ($f_{\mathrm{CE}}$).
The IMU factor enforces motion consistency between consecutive states by constraining the evolution of pose and velocity through preintegrated inertial measurements.
The radar ego-velocity factor includes direct velocity observations whose residual explicitly depends on the radar–IMU extrinsic transformation and temporal offset, allowing both calibration parameters to be corrected online as part of the state optimization. 
The temporal offset consistency factor and extrinsic calibration consistency factor penalize abrupt changes in the respective calibration parameters between consecutive states, ensuring smooth online convergence.

\subsection{Continuous-Time Inertial Modeling via Uniform Cubic B-Splines}

One of our contributions lies in refining the radar factor through continuous-time inertial modeling using uniform cubic B-splines \cite{bartels1995introduction}.
Given $N$ discrete IMU measurements $\mathcal{D} = \{(t_j, \mathbf{a}_j,\boldsymbol{\omega}_j)\}_{j=1}^{N}$ within a window centered on the radar timestamp, we fit independent uniform cubic B-splines to the raw acceleration and angular velocity signals.
A uniform knot vector spanning $[t_{\min}, t_{\max}]$ with $n_s$ segments of width $h = (t_{\max} - t_{\min}) / n_s$ yields $n_c = n_s + 3$ control points $\{\mathbf{c}_i\}$, and the spline is evaluated on segment $i$ as

\begin{equation}
\begin{aligned}
S(t) &= \mathbf{a}_0 + \mathbf{a}_1 u + \mathbf{a}_2 u^2 + \mathbf{a}_3 u^3, \\
u &= \frac{t - (t_{\min} + i \cdot h)}{h} \in [0,1),
\end{aligned}
\label{eq:bspline_eval}
\end{equation}
where the polynomial coefficients follow from the four local control points via the standard uniform cubic B-spline basis matrix:
\begin{equation}
\begin{bmatrix} \mathbf{a}_0 \\ \mathbf{a}_1 \\ \mathbf{a}_2 \\ \mathbf{a}_3
\end{bmatrix}
=
\frac{1}{6}
\begin{bmatrix}
 1 &  4 &  1 & 0 \\
-3 &  0 &  3 & 0 \\
 3 & -6 &  3 & 0 \\
-1 &  3 & -3 & 1
\end{bmatrix}
\begin{bmatrix} \mathbf{c}_i \\ \mathbf{c}_{i+1} \\ \mathbf{c}_{i+2} \\
\mathbf{c}_{i+3} \end{bmatrix}.
\label{eq:bspline_basis}
\end{equation}
The control points are obtained by solving the regularized least-squares
problem
\begin{equation}
\min_{\{\mathbf{c}_i\}} \sum_{j=1}^{N}
\left\| \mathbf{y}_j - S(t_j) \right\|^2
+ \lambda \sum_{i} \left\| \mathbf{c}_i \right\|^2,
\label{eq:bspline_fit}
\end{equation}
where $\lambda$ is a Tikhonov regularization parameter that prevents overfitting. 
To handle negative temporal offsets (i.e., the IMU lags behind the radar), synthetic measurements replicating the last known acceleration and angular velocity are appended beyond the data-supported domain, effectively extending the spline with a constant extrapolation corresponding to the most recent observed inertial state.

\subsection{Radar Ego-Velocity Factor with Temporal Compensation}

Using the continuous-time inertial model, we define the radar ego-velocity 
measurement function as
\begin{equation}
\begin{aligned}
h^R(\mathbf{x}) =
\mathbf{R}_{\mathrm{RI}}\Big(
&\Delta\mathbf{R}(t,\, t + t_{O}) \,\mathbf{R}_{\mathrm{IW}}\,  
{}_{\!W}\mathbf{v} \\
&+ \boldsymbol{\omega}^{BS}(t + t_{O}) \times {}_{\!I}\mathbf{t}_{\mathrm{RI}}\\
&+ \Delta \mathbf{v}(t,\, t + t_{O})
\Big),
\end{aligned}
\label{eq:radar_factor}
\end{equation}
where $\boldsymbol{\omega}^{BS}$ and $\mathbf{a}^{BS}$ denote the  B-spline posterior evaluations of angular velocity and linear acceleration,  respectively.

The velocity increment is computed by integrating the B-spline acceleration  model over the temporal offset interval,
\begin{equation}
\Delta \mathbf{v}(t, t+t_O)
=
\int_{t}^{t+t_O}
\left( \mathbf{a}^{BS}(s) - \mathbf{b}_a - \mathbf{R}_{\mathrm{IW}}(s) \mathbf{g} \right)
\mathrm{d}s,
\label{eq:bs_dv}
\end{equation}
where $\mathbf{g}$ is the gravity vector and $\mathbf{b}_a$ is the  accelerometer bias.
The integral is evaluated using 5-point Gauss-Legendre quadrature, mapping  the integration variable as $s = t+t_O \cdot \xi$ for $\xi \in [0,1]$:
\begin{equation}
\Delta \mathbf{v} \approx t_O \sum_{i=1}^{5} w_i\, 
\tilde{\mathbf{a}}^{BS}(s_i),
\label{eq:quadrature}
\end{equation}
where $\xi_i$ and $w_i$ are the 5-point Gauss-Legendre nodes and  weights, and $\tilde{\mathbf{a}}^{BS}$ denotes the specific force in IMU frame after gravity and bias subtraction.

The relative rotation over the offset interval is obtained by integrating  the B-spline angular velocity, where $\exp(\cdot)_{\times}$ denotes the 
Lie group exponential map $\mathfrak{so}(3) \rightarrow SO(3)$ mapping 
body-frame angular velocities to rotation matrices,
\begin{equation}
\Delta\mathbf{R}(t,\, t+t_O)
=
\exp\!\left(
  \int_{t}^{t+t_O}
   (\boldsymbol{\omega}^{BS}(s) - \mathbf{b}_g)\, \mathrm{d}s
\right)_{\!\times},
\label{eq:bs_dR}
\end{equation}
approximated at each quadrature point. 
Therefore, $\mathbf{R}_{\mathrm{IW}}$ rotation is approximated  at each quadrature point $s_i$ as $\mathbf{R}_{\mathrm{IW}}(s_i) \approx \exp\!\left(\boldsymbol{(\omega}^{BS}(s_i)-\mathbf{b}_g)\,(s_i - t)\right)_{\!\times}\mathbf{R}_{\mathrm{IW}}(t)$, which rotates the gravity vector into the IMU body frame at each  integration point.

The lever-arm term $\boldsymbol{\omega}^{BS}(t + t_O) \times  {}_I\mathbf{t}_{\mathrm{RI}}$ models the additional linear velocity at the  radar phase center induced by the instantaneous angular rate.

\subsection{Temporal Offset, Extrinsic Calibration, and IMU Factors}

To ensure smooth temporal evolution of the estimated offset, a  constant-time consistency factor is introduced:
\begin{equation}
h^{\mathrm{CT}}(\mathbf{x}_{k-1}) = t_{O,k-1}.
\end{equation}
This factor enforces a slowly varying temporal offset by penalizing abrupt  changes between consecutive states.

Similarly, to maintain physically plausible radar–IMU geometry during online  calibration, a constant extrinsic consistency factor is employed:
\begin{equation}
h^{\mathrm{CE}}(\mathbf{x}_{k-1}) =
\begin{bmatrix}
\mathbf{R}_{\mathrm{RI},k-1} \\
{}_{\!I}\mathbf{p}_{\mathrm{RI},k-1}
\end{bmatrix}.
\end{equation}
This factor discourages rapid variations in the estimated extrinsic parameters  while still allowing gradual corrections driven by radar measurements.
The IMU preintegration factor follows standard  formulations~\cite{imu_factor,mfi-rik}, relating consecutive states through  preintegrated inertial measurements.

\subsection{Optimization and Implementation}

The complete optimization problem is formulated as
\begin{equation}
\begin{aligned}
\hat{\mathbf{x}} =
\arg\min_{\mathbf{x}}
\sum_k
\Big(
\| \mathbf{e}^{\mathrm{IMU}}_k \|^2_{\Sigma^{\mathrm{IMU}}_k}
+
\rho\!\left(
\| \mathbf{e}^{R}_k \|^2_{\Sigma^{R}_k}
\right) \\
+
\| \mathbf{e}^{\mathrm{CT}}_k \|^2_{\Sigma^{\mathrm{CT}}_k}
+
\| \mathbf{e}^{\mathrm{CE}}_k \|^2_{\Sigma^{\mathrm{CE}}_k}
\Big),
\end{aligned}
\end{equation}
where $\rho(\cdot)$ denotes a Huber loss for robustness to radar 
ego-velocity outliers.

The B-spline fit is computed once per radar measurement over a local IMU window spanning 140 ms into the past and 20 ms into the future, sized to encompass the range of temporal offsets commonly encountered in RIO systems \cite{kim2025ekf, vstironja2025impact}.
This window is a tunable parameter that can be adjusted based on the expected offset range of a given sensor configuration.
The system is implemented using GTSAM~\cite{gtsam_perception} with iSAM2 within a one-second sliding window. 
A two-thread architecture is employed, an optimization thread solving the MAP problem over the sliding window, and a navigation thread propagating the latest state estimate at the IMU rate. 
All odometry results reported in Section~\ref{sec:results} are obtained from the navigation thread output to ensure fair comparison.

\section{Experimental Results}
\label{sec:results}

We evaluate the proposed method, LC-RIO-ET, on two publicly available radar-inertial datasets: the EKF-RIO-TC dataset~\cite{kim2025ekf} and the ICINS dataset~\cite{doer2021yaw}.
These datasets provide complementary evaluation conditions: the EKF-RIO-TC dataset lacks hardware synchronization between radar and IMU, making it well-suited for evaluating online temporal calibration, whereas the ICINS dataset employs a microcontroller-based hardware trigger to ensure accurate sensor synchronization.
This distinction allows us to assess the benefit of online spatio-temporal calibration in both synchronized and unsynchronized settings.

The EKF-RIO-TC dataset was collected using a Texas Instruments AWR1843BOOST radar and an Xsens MTi-670 IMU, with ground truth trajectories obtained via a motion capture system.
The ICINS dataset employs a TI IWR6843AOP radar, an ADIS16448 IMU, and a monocular camera. Since motion capture is unavailable, ground truth is obtained via visual--inertial SLAM with multiple loop closures and is therefore treated as pseudo ground truth.

We compare LC-RIO-ET against three external baselines: the standard EKF-based estimator EKF-RIO~\cite{doer_ekf}, state-of-the-art EKF-based RIO with online temporal calibration EKF-RIO-TC~\cite{kim2025ekf}, and the FGO-based RIO with online temporal calibration RIO-T~\cite{vstironja2025impact}.
In addition, we evaluate four ablation variants of our formulation: (i) LC-RIO, a plain FGO without temporal or extrinsic calibration, (ii) LC-RIO-E, with online extrinsic calibration, (iii) LC-RIO-T, with online temporal calibration, and (iv) LC-RIO-ET, the proposed method with joint online spatio-temporal calibration.

All results are evaluated using the EVO toolbox~\cite{evo}.
We report the Absolute Pose Error (APE) under $\mathrm{SE}(3)$ alignment and the Relative Pose Error (RPE) computed over 10\,m intervals.
RPE captures short-term motion consistency critical for local navigation, while APE quantifies long-term drift.
Each method is executed five times and we report the mean across runs to reduce stochastic effects.
For all methods performing online temporal calibration, the temporal offset is initialized to zero. Gyroscope bias is initialized using a 12\,s stationary phase.
All methods receive the same input data, with radar ego-velocity estimates computed using the \textit{reve} package~\cite{doer_ekf}.

We evaluate three aspects: (1) overall odometry accuracy on both datasets, (2) the impact of spatio-temporal calibration on downstream odometry when calibrated parameters are re-injected as initial values into baseline estimators, and (3) convergence of the estimated temporal offset.

\begin{table}[t]
\vspace{5pt}
\centering
\caption{Quantitative comparison on the EKF-RIO-TC dataset~\cite{kim2025ekf}.}
\setlength{\tabcolsep}{3pt}
\renewcommand{\arraystretch}{1.2}
\begin{tabular}{l l cc cc}
\hline
\textbf{Sequence} & \textbf{Method} & \multicolumn{2}{c}{\textbf{APE Mean}} & \multicolumn{2}{c}{\textbf{RPE Mean}} \\
 & & Trans. (m) & Rot. ($^\circ$) & Trans. (m) & Rot. ($^\circ$) \\
\hline

\multirow{2}{*}{1}
& EKF-RIO & 0.583 & 2.997 & 0.354 & 2.707 \\
& EKF-RIO-TC  & 0.333 & 1.907 & 0.175 & 2.412 \\
& RIO-T & 0.435 & 2.102 & 0.183 & 2.618 \\
\cdashline{2-6}
& LC-RIO & 0.411 & 1.915 & 0.196 & 2.992 \\
& LC-RIO-E & 0.586 & 1.794 & 0.224 & 3.081 \\
& LC-RIO-T & 0.380 & 1.614 & 0.171 & 2.181 \\
& LC-RIO-ET & \textbf{0.324} & \textbf{1.570} & \textbf{0.161} & \textbf{2.131} \\
\hline

\multirow{2}{*}{2}
& EKF-RIO & 0.870 & 6.150 & 0.306 & 2.684 \\
& EKF-RIO-TC & 0.372 & 2.323 & 0.153 & 2.582 \\
& RIO-T & 0.412 & 2.046 & 0.175 & 2.238 \\
\cdashline{2-6}
& LC-RIO & 0.379 & 1.996 & 0.173 & \textbf{1.890} \\
& LC-RIO-E & 0.376 & 2.078 & 0.171 & 1.997 \\
& LC-RIO-T & 0.278 & \textbf{1.967} & 0.143 & 2.324 \\
& LC-RIO-ET & \textbf{0.219} & 1.992 & \textbf{0.140} & 2.284 \\
\hline

\multirow{2}{*}{3}
& EKF-RIO & 0.772 & 3.619 & 0.314 & 2.299 \\
& EKF-RIO-TC & 0.216 & 2.291 & 0.131 & 1.876 \\
& RIO-T & 0.720 & 3.915 & 0.205 & 2.226 \\
\cdashline{2-6}
& LC-RIO & 0.657 & 3.769 & 0.199 & 2.033 \\
& LC-RIO-E & 0.719 & 4.885 & 0.209 & 2.037 \\
& LC-RIO-T & \textbf{0.206} & 2.003 & 0.131 & 1.977 \\
& LC-RIO-ET & 0.219 & \textbf{1.985} & \textbf{0.127} & \textbf{1.861} \\
\hline

\multirow{2}{*}{4}
& EKF-RIO & 0.827 & 16.523 & 0.316 & 4.844 \\
& EKF-RIO-TC & 0.264 & \textbf{2.730} & 0.167 & \textbf{2.991} \\
& RIO-T & 0.268 & 3.764 & 0.221 & 3.214 \\
\cdashline{2-6}
& LC-RIO & \textbf{0.247} & 3.362 & 0.176 & 3.156 \\
& LC-RIO-E & 0.329 & 3.281 & 0.138 & 3.279 \\
& LC-RIO-T & 0.320 & 3.253 & 0.183 & 3.002 \\
& LC-RIO-ET & 0.304 & 3.074 & \textbf{0.116} & 3.114 \\
\hline

\multirow{2}{*}{5}
& EKF-RIO & 1.311 & 6.122 & 0.386 & 3.391 \\
& EKF-RIO-TC & 0.570 & 2.601 & 0.221 & 2.591 \\
& RIO-T & 0.614 & 2.958 & 0.264 & 3.337 \\
\cdashline{2-6}
& LC-RIO & 0.713 & 2.812 & 0.287 & 3.584 \\
& LC-RIO-E & 0.378 & 1.948 & 0.174 & 3.061 \\
& LC-RIO-T & 0.620 & 2.500 & 0.222 & 2.548 \\
& LC-RIO-ET & \textbf{0.284} & \textbf{1.693} & \textbf{0.146} & \textbf{2.494} \\
\hline
\hline

\textbf{Mean}
& EKF-RIO \cite{doer_ekf} & 0.873 & 7.082 & 0.335 & 3.185 \\
& EKF-RIO-TC \cite{kim2025ekf} & 0.351 & 2.370 & 0.169 & 2.490 \\
& RIO-T \cite{vstironja2025impact} &  0.490 & 2.957 & 0.210 & 2.726 \\
\cdashline{2-6}
& LC-RIO & 0.481 & 2.771 & 0.206 & 2.731 \\
& LC-RIO-E & 0.477 & 2.797 & 0.183 & 2.691 \\
& LC-RIO-T & 0.361 & 2.267 & 0.170 & 2.406 \\
& LC-RIO-ET & \textbf{0.270} & \textbf{2.063} & \textbf{0.138} & \textbf{2.377} \\
\hline
\end{tabular}
\label{odometry1}
\vspace{-4mm}
\end{table}

\begin{table}[t]
\vspace{5pt}
\centering
\caption{Quantitative comparison on the ICINS dataset~\cite{doer2021yaw}.}
\setlength{\tabcolsep}{3pt}
\renewcommand{\arraystretch}{1.2}
\begin{tabular}{l l cc cc}
\hline
\textbf{Sequence} & \textbf{Method} & \multicolumn{2}{c}{\textbf{APE Mean}} & \multicolumn{2}{c}{\textbf{RPE Mean}} \\
 & & Trans. (m) & Rot. ($^\circ$) & Trans. (m) & Rot. ($^\circ$) \\
\hline

\multirow{2}{*}{carried 1}
& EKF-RIO & 0.995 & 5.553 & 0.134 & 0.857 \\
& EKF-RIO-TC & 0.922 & 5.102 & 0.139 & 0.798 \\
& RIO-T & 0.754 & 4.098 & \textbf{0.126} & \textbf{0.641} \\
\cdashline{2-6}
& LC-RIO & \textbf{0.745} & \textbf{4.037} & 0.128 & 0.655 \\
& LC-RIO-E & 0.863 & 5.228 & 0.141 & 0.687 \\
& LC-RIO-T & 0.763 & 4.053 & 0.128 & 0.665 \\
& LC-RIO-ET & 0.768 & 4.038 & \textbf{0.126} & 0.666 \\
\hline

\multirow{2}{*}{carried 2}
& EKF-RIO & 1.761 & 10.743 & 0.150 & 1.083 \\
& EKF-RIO-TC & 1.594 & 9.613 & 0.152 & 0.992 \\
& RIO-T & 1.519 & 9.258 & \textbf{0.126} & \textbf{0.841} \\
\cdashline{2-6}
& LC-RIO & \textbf{1.513} & 9.094 & \textbf{0.126} & 0.845 \\
& LC-RIO-E & 2.091 & 12.385 & 0.182 & 0.886 \\
& LC-RIO-T & 1.526 & 9.095 & 0.128 & 0.863 \\
& LC-RIO-ET & 1.518 & \textbf{9.071} & 0.128 & 0.871 \\
\hline

\multirow{2}{*}{flight 1}
& EKF-RIO & 0.736 & 4.944 & 0.213 & \textbf{1.401} \\
& EKF-RIO-TC & 0.764 & 5.274 & 0.219 & 1.493 \\
& RIO-T & 0.699 & 4.725 & 0.189 & 1.428 \\
\cdashline{2-6}
& LC-RIO & 0.699 & 4.783 & 0.191 & 1.424 \\
& LC-RIO-E & 0.698 & 4.713 & \textbf{0.180} & 1.424 \\
& LC-RIO-T & \textbf{0.695} & 4.727 & 0.189 & 1.419 \\
& LC-RIO-ET & 0.698 & \textbf{4.685} & 0.190 & 1.423 \\
\hline

\multirow{2}{*}{flight 2}
& EKF-RIO & 0.100 & 1.938 & 0.176 & 0.888 \\
& EKF-RIO-TC & 0.098 & 1.815 & 0.161 & 0.914 \\
& RIO-T & 0.098 & 1.727 & \textbf{0.153} & 0.817 \\
\cdashline{2-6}
& LC-RIO & 0.101 & 1.759 & 0.157 & 0.836 \\
& LC-RIO-E & 0.102 & 1.725 & 0.160 & 0.796 \\
& LC-RIO-T & 0.099 & 1.720 & 0.155 & \textbf{0.792} \\
& LC-RIO-ET & \textbf{0.096} & \textbf{1.663} & 0.160 & 0.805 \\
\hline
\hline
\textbf{Mean}
& EKF-RIO \cite{doer_ekf} & 0.898 & 5.795 & 0.168 & 1.057 \\
& EKF-RIO-TC \cite{kim2025ekf}& 0.845 & 5.451 & 0.168 & 1.049 \\
& RIO-T \cite{vstironja2025impact} & 0.768 & 4.952 & \textbf{0.149} & \textbf{0.932} \\
\cdashline{2-6}
& LC-RIO & \textbf{0.764} & 4.918 & 0.151 & 0.940 \\
& LC-RIO-E & 0.939 & 6.013 & 0.166 & 0.948 \\
& LC-RIO-T & 0.771 & 4.899 & 0.150 & 0.935 \\
& LC-RIO-ET & 0.770 & \textbf{4.864} & 0.151 & 0.941 \\
\hline
\end{tabular}
\label{table2}
\vspace{-4mm}
\end{table}

%-----------------------------------------------------------------------
\subsection{Odometry Accuracy}
%-----------------------------------------------------------------------

\paragraph{EKF-RIO-TC dataset (unsynchronized).}
Table~\ref{odometry1} reports per-sequence and mean results on the EKF-RIO-TC dataset.
LC-RIO-ET achieves the best mean performance across all four metrics, with a mean translational RPE of 0.138\,m and rotational RPE of 2.377$^\circ$, representing improvements of 18.3\% and 4.5\%, respectively, over EKF-RIO-TC.
Relative to the plain FGO baseline LC-RIO, the translational RPE improves by 33.0\% and the translational APE by 43.9\%, confirming that jointly estimating both calibration parameters is substantially more beneficial than either alone.

The ablation study reveals complementary roles for temporal and extrinsic calibration.
LC-RIO-T alone reduces mean translational RPE by 17.5\% relative to LC-RIO, demonstrating that temporal compensation is the dominant contributor in the unsynchronized setting.
LC-RIO-E yields only marginal improvements in isolation, as extrinsic calibration requires sufficient rotational excitation to converge reliably.
The joint LC-RIO-ET formulation consistently outperforms both individual strategies when excitation is sufficient, as most clearly evidenced in Sequences 4 and 5, which feature the richest rotational motion, e.g., in Sequence 5, LC-RIO-ET reduces translational RPE by 49.1\% relative to LC-RIO.

Compared against RIO-T, which is also implemented within a factor-graph framework but relies on locally constant acceleration assumptions, LC-RIO-ET reduces mean translational RPE by 34.3\%.
Due to this local assumption, RIO-T requires several iterations to converge to the temporal offset, whereas the continuous-time B-spline model enable reliable temporal offset estimation even under large offsets.

\paragraph{ICINS dataset (hardware-synchronized).}
Table~\ref{table2} reports results on the ICINS dataset, where the hardware trigger ensures a smaller temporal offset. 
Unlike the EKF-RIO-TC dataset, where the estimated offset converges to approximately 113\,ms, the estimated offset on this dataset converges to approximately -10\,ms, which under the adopted sign convention \cite{vstironja2025impact} indicates that the IMU measurements arrive slightly after the radar measurements.
LC-RIO-ET performs on par with LC-RIO, with nearly identical mean translational RPE, confirming that the method does not degrade performance in hardware-synchronized settings with limited motion dynamics. 
Notably, LC-RIO-ET achieves the lowest mean rotational APE (4.864$^\circ$), suggesting a residual benefit of the continuous-time inertial model even in synchronized conditions. 
The degradation observed for LC-RIO-E on the carried sequences confirms that extrinsic calibration without sufficient rotational observability can introduce errors rather than correct it. 
Importantly, even where immediate odometry gains are limited, LC-RIO-ET reliably estimates spatio-temporal calibration parameters that directly benefit any downstream task, as demonstrated in Section~\ref{sec:impact}.

\begin{table}[t]
\vspace{5mm}
\centering
\caption{Impact of online spatio-temporal calibration: performance of
EKF-RIO and LC-RIO when re-initialized with extrinsic and temporal
parameters estimated by LC-RIO-ET (denoted by $^*$). Best results per
column in bold.}
\setlength{\tabcolsep}{4pt}
\renewcommand{\arraystretch}{1.2}
\begin{tabular}{l l cc cc}
\hline
\textbf{Sequence} & \textbf{Method} & \multicolumn{2}{c}{\textbf{APE Mean}} & \multicolumn{2}{c}{\textbf{RPE Mean}} \\
 & & Trans. (m) & Rot. ($^\circ$) & Trans. (m) & Rot. ($^\circ$) \\
\hline

\multirow{2}{*}{5}
& EKF-RIO & 1.311 & 6.122 & 0.386 & 3.391 \\
& EKF-RIO$^*$ & \textbf{0.255} & \textbf{1.643} & \textbf{0.122} & \textbf{2.409} \\
\cdashline{2-6}
& LC-RIO & 0.713 & 2.812 & 0.287 & 3.584 \\
& LC-RIO$^*$ & \textbf{0.293} & \textbf{1.607} & \textbf{0.140} & \textbf{2.276} \\
\hline

\multirow{2}{*}{flight 2}
& EKF-RIO & \textbf{0.100} & 1.938 & 0.176 & 0.888 \\
& EKF-RIO$^*$ &  0.112  & \textbf{1.783} & \textbf{0.165} & \textbf{0.867} \\
\cdashline{2-6}
& LC-RIO  & \textbf{0.101} & 1.759 & 0.157 & 0.836 \\
& LC-RIO$^*$ & 0.102  & \textbf{1.714} & \textbf{0.154} & \textbf{0.835} \\
\hline

\end{tabular}
\label{tab:impact}
\end{table}

\begin{figure}[t]
    \centering
    \includegraphics[width=0.4\textwidth]{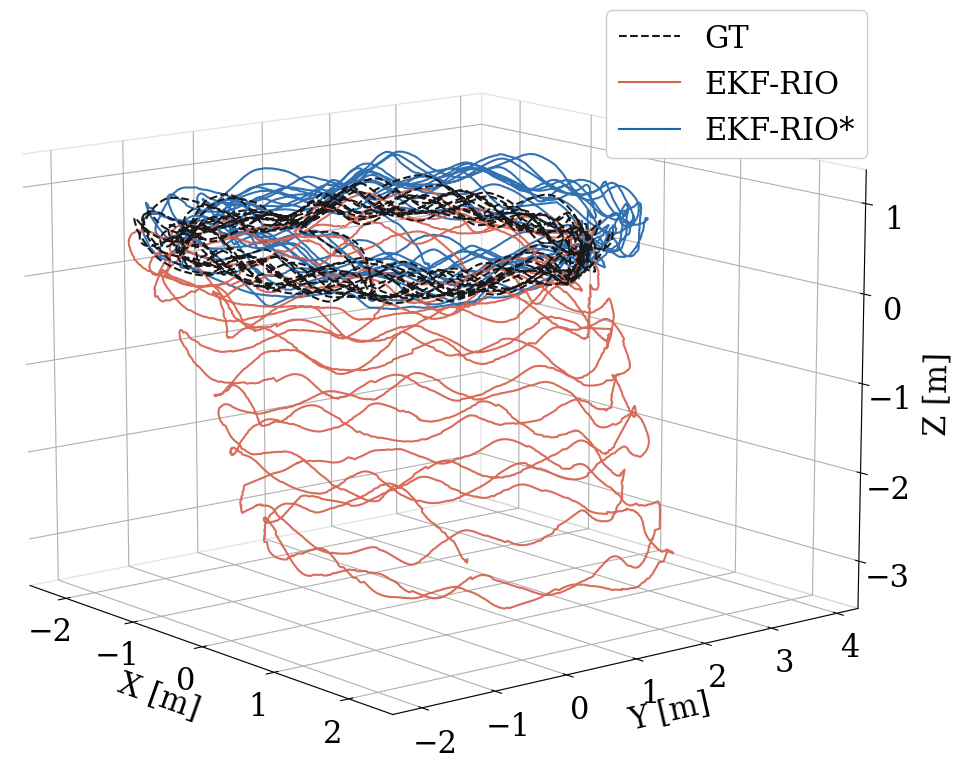}
    \caption{Trajectory visualization of EKF-RIO with original and estimated parameters on Sequence 5 from EKF-RIO-TC dataset~\cite{kim2025ekf}.}
    \label{fig::3dviz}
\end{figure}

\begin{figure}[t]
    \centering
    \begin{subfigure}[t]{\columnwidth}
        \centering
        \includegraphics[width=0.98\linewidth]{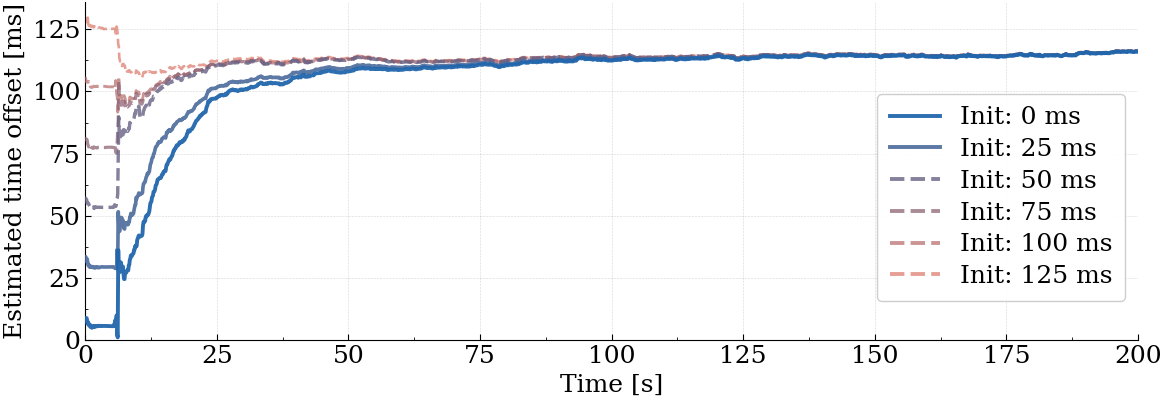}
        \caption{Temporal offset convergence under various initial values.}
        \label{fig::t_conv}
    \end{subfigure}
    \\[6pt]
    \begin{subfigure}[t]{\columnwidth}
        \centering
        \includegraphics[width=0.98\linewidth]{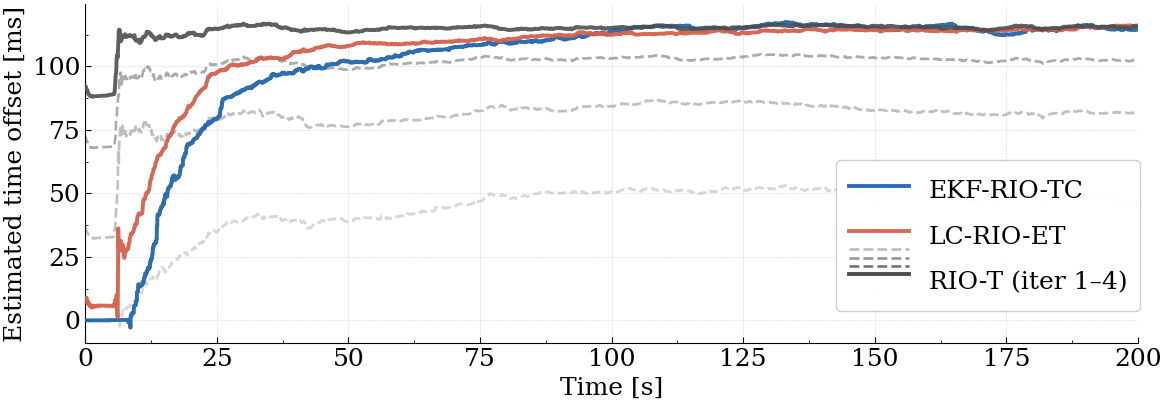}
        \caption{Temporal offset of LC-RIO-ET, EKF-RIO-TC, and RIO-T.}
        \label{fig::met}
    \end{subfigure}
    \caption{Temporal offset estimation results. (a) Convergence under various
    initial values. (b) Comparison of LC-RIO-ET, EKF-RIO-TC, and RIO-T.}
    \label{fig::temporal_offset}
\end{figure}

\subsection{Impact of Online Spatio-Temporal Calibration}
\label{sec:impact}

To isolate the value of the calibrated parameters independently of the FGO formulation itself, we reinitialize EKF-RIO and LC-RIO with the extrinsic and temporal offset parameters estimated by LC-RIO-ET (denoted by $^*$), and re-run both estimators with these fixed calibration values.
Results are reported in Table~\ref{tab:impact} for Sequence 5 (unsynchronized, high dynamics) and \textit{flight~2} (synchronized).

On Sequence 5, the improvements are substantial across all metrics, calibrated parameters reduce translational RPE of EKF-RIO by 68.4\% and of LC-RIO by 51.2\%, with corresponding translational APE reductions of 80.5\% and 58.9\% (trajectories visualized in Fig.~\ref{fig::3dviz}).
On the synchronized \textit{flight~2} sequence, re-injecting  calibrated parameters consistently reduces rotational errors and  translational RPE, while translational APE  increases marginally.
This is particularly significant in the context of the ICINS results, where LC-RIO-ET did not outperform LC-RIO in online odometry, even in that synchronized setting, the converged calibration parameters significantly enhance the accuracy of both baseline estimators, confirming that the value of spatio-temporal calibration extends beyond the online estimation window.

\subsection{Temporal Offset Accuracy and Convergence}

Figure~\ref{fig::temporal_offset} summarizes the temporal offset estimation behavior of LC-RIO-ET.

\paragraph{Convergence from diverse initializations.}
Figure~\ref{fig::t_conv} shows the temporal offset trajectory of LC-RIO-ET on Sequence 5 initialized from six values: 0, 25, 50, 75, 100, and 125\,ms.
In all cases the estimate converges robustly to a consistent steady-state value of approximately 113.3\,ms, demonstrating that the B-spline-based radar factor provides a well-conditioned gradient signal with respect to the temporal offset across a wide initialization range.

\paragraph{Comparison with competing methods.}
Figure~\ref{fig::met} compares the temporal offset trajectories of LC-RIO-ET, EKF-RIO-TC, and RIO-T on Sequence 5.
EKF-RIO-TC converges to 115.5\,ms, RIO-T converges in 4 iterations to 118.0\,ms, and LC-RIO-ET converges to 113.6\,ms, providing cross-method validation of the underlying temporal offset.
Due to its locally constant acceleration assumption, RIO-T requires multiple iterations to resolve large temporal offsets, whereas LC-RIO-ET achieves convergence online even for larger temporal offsets due to continuous time IMU modeling.
A key distinction is that LC-RIO-ET simultaneously converges the extrinsic calibration parameters alongside the temporal offset, whereas EKF-RIO-TC and RIO-T assume fixed extrinsics throughout.

\subsection{Discussion}

Online temporal calibration is the dominant performance driver in the unsynchronized EKF-RIO-TC dataset, with LC-RIO-T alone recovering the majority of the accuracy gap over LC-RIO. 
The continuous-time B-spline inertial model further improves over the locally constant acceleration assumption of RIO-T under high-dynamics motion, enabling reliable online convergence even for large temporal offsets directly within the FGO, without requiring iterative refinement.
Extrinsic calibration yields consistent gains when rotational excitation is sufficient for observability, but can introduce drift otherwise, suggesting that conditioning the extrinsic calibration factor on an observability criterion would improve robustness in low-excitation scenarios.

On the hardware-synchronized ICINS dataset, LC-RIO-ET performs on par with the baselines, confirming that the method does not degrade performance when temporal offsets are negligible. 
The calibration parameter injection experiment further underscores the practical value of the proposed approach: spatio-temporal calibration parameters estimated by LC-RIO-ET transfer directly to simpler estimators, yielding substantial RPE decrease.

Looking ahead, several directions present promising extensions.
First, integrating place recognition, e.g., scan-context-based place recognition~\cite{kim2018scan} could enable loop-closure correction and globally consistent mapping, particularly relevant for long-duration missions.
Second, extending to a tightly coupled formulation that directly processes raw radar point clouds could improve robustness in highly dynamic environments where the static-scene assumption underlying ego-velocity estimation can be violated.
Third, adaptive tuning of the B-spline segment count $n_s$ and regularization parameter $\lambda$ based on detected motion dynamics could further improve offset estimation under variable-speed trajectories.

\section{Conclusion}
\label{sec:conclusion}

In this paper we have presented a loosely coupled radar-inertial odometry system with online joint spatio-temporal calibration within a factor graph framework.
The core contribution is a continuous-time inertial model based on uniform cubic B-splines embedded directly in the radar ego-velocity factor, enabling simultaneous online estimation of the radar-IMU temporal offset and extrinsic calibration without assuming locally constant acceleration or hardware synchronization.
On the unsynchronized dataset, our proposed method achieves the best performance across all metrics, reducing mean translational RPE by 18.3\% over the strongest baseline and by 33.0\% over a plain FGO without calibration, while performing on par with the best methods on the hardware-synchronized ICINS dataset.
Injecting the estimated calibration parameters into standard baseline estimators yields RPE reductions of up to 68\%, underscoring the practical value of accurate spatio-temporal calibration as a standalone module.

Future work will explore loop-closure integration, tight coupling with raw radar point clouds, and adaptive B-spline parameterization.

\addtolength{\textheight}{-5cm}   % This command serves to balance the column lengths

\section*{ACKNOWLEDGMENT}
This research has been supported by the European Regional Development Fund under grant agreement PK.1.1.10.0007 (DATACROSS) and the Croatian Science Foundation under grant agreement DOK-NPOO-2023-10-6705.

%%%%%%%%%%%%%%%%%%%%%%%%%%%%%%%%%%%%%%%%%%%%%%%%%%%%%%%%%%%%%%%%%%%%%%%%%%%%%%%%

\balance
\bibliography{root}
\bibliographystyle{IEEEtran}

\end{document}